\pdfoutput=1

\documentclass[11pt]{article}

\usepackage[]{EMNLP2023}

\usepackage{times}
\usepackage{latexsym}

\usepackage[T1]{fontenc}

\usepackage[utf8]{inputenc}

\usepackage{microtype}
\usepackage{booktabs}       
\usepackage{makecell}
\usepackage{amsmath}

\colorlet{linecol}{black!75}
\usepackage{forest}
\usepackage{cuted}

\usepackage{cleveref}
\crefname{section}{§}{§§}
\Crefname{section}{§}{§§}
\usepackage{inconsolata}

\title{Towards Better Chain-of-Thought Prompting Strategies:\\ A Survey}

\author{Zihan Yu \quad {\bf Liang He} \quad  {\bf Zhen Wu} \thanks{*Corresponding author} \quad {\bf Xinyu Dai}  \quad {\bf Jiajun Chen}\\
         National Key Laboratory for Novel Software Technology, Nanjing University, China\\
         Collaborative Innovation Center of Novel Software Technology and Industrialization, China\\
         \texttt{\{zihan.y, heliang\}@smail.nju.edu.cn}\\
         \texttt{\{wuz, daixinyu, chenjj\}@nju.edu.cn}}

\begin{document}
\maketitle
\begin{abstract}
Chain-of-Thought (CoT), a step-wise and coherent reasoning chain, shows its impressive strength when used as a prompting strategy for large language models (LLM). Recent years, the prominent effect of CoT prompting has attracted emerging research. However, there still lacks of a systematic summary about key factors of CoT prompting and comprehensive guide for prompts utilizing. For a deeper understanding about CoT prompting, we survey on a wide range of current research, presenting a systematic and comprehensive analysis on several factors that may influence the effect of CoT prompting, and introduce how to better apply it in different applications under these discussions. We further analyze the challenges and propose some future directions about CoT prompting. This survey could provide an overall reference on related research.
\end{abstract}

\section{Introduction}
Recent years, large language models (LLM) with prompting strategies have achieved prominent performance on many traditional NLP benchmarks \cite{GPT3, cui-etal-2021-template, palm, flan-palm, li2022selfprompting}. But some work finds vanilla prompting strategies still struggle to improve LLM performance on multi-step tasks \cite{CoT, zeroshot-CoT, faithful-CoT, synthetic}. 

Based on natural step-by-step thinking ability of humans, Chain-of-Thought (CoT) prompting is proposed for LLM to solve multi-step reasoning problems \cite{CoT}. This prompting strategy tries to incorporate intermediate steps to guide a progressive reasoning, achieving surprising improvement on many reasoning benchmarks \cite{CoT, selfconsistency} even in some special scenarios including cross-domain \cite{huang2022large}, length-generalization \cite{anil2022exploring, LtM} and cross-lingual reasoning \cite{shi2022language}. Additionally, CoT prompting ensures a logical and traceable reasoning process, which is more interpretable for humans \cite{maieutic, weng2023large}.  

The impressive result of CoT prompting provokes an upsurge of explorations on CoT prompting strategies design across different tasks on different models \cite{synthetic, selfconsistency, decomposed, LtM}. But before  designing a specific strategy, it's necessary to systematically understand what factors may influence the performance of CoT prompting. Besides, although CoT prompting shows its advantages in reasoning tasks, it still has some limitations on generality and concerning on its opaque mechanism as well as unfaithful generation \cite{faithful-CoT, shi2022language, suzgun2022challenging, zhang2023multimodal}. To better guide further work, it is essential to make a discussion on current challenges of CoT. Existing surveys lack of a specialized formulation as well as a deep analysis on CoT prompting. We discuss existing surveys and compare them to our work in Appendix \ref{A}.

\begin{figure*}
\centering
\includegraphics[height=6.75cm,width=15cm]{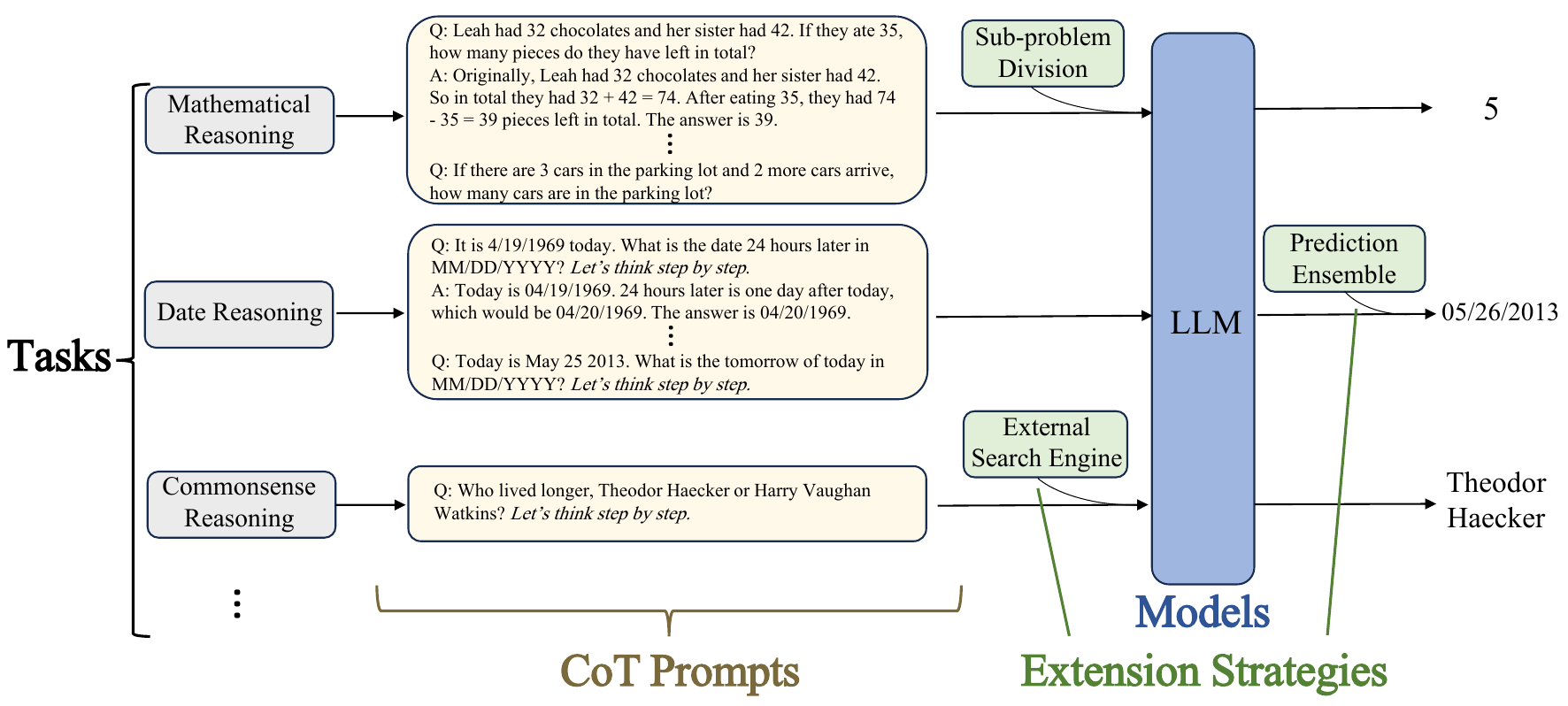}  
\caption{Illustration of the general pipeline to utilize Chain-of-Thought prompting strategy. Given a specific task, CoT prompts are designed on it. With the assistance of optional extension strategies, models predict the answers (typically with CoT rationales) based on the input prompts.}
\label{overview}
\end{figure*}

This survey presents a comprehensive and systematic analysis of the CoT prompting effect, making several noteworthy contributions to this field. First, we formalize the CoT prompting and summarize the pipeline when using it (as shown in Figure \ref{overview}), which forms the basis of our analysis and discussions. Then, we identify four key factors that significantly affect the performance of CoT prompting: task types, prompts design, extension strategies, and models. We conduct a detailed exploration on how these factors impact the prompting effect and introduce innovative methods inspired by them. Moreover, we provide a comprehensive vision of current challenges of CoT prompting and outline numerous opportunities for future research.

\paragraph{Overview of this survey:} We first present some background knowledge in \cref{2} then make an introduction of CoT prompting in \cref{3}. Subsequently, we discuss several key factors about CoT prompting including tasks (\cref{4}), prompts (\cref{7}), extension strategies (\cref{6}) and models (\cref{5}). We further discuss the challenges and future directions in \cref{8}.

\section{Background Knowledge}
\label{2}
In this part, we will make a brief introduction about background knowledge before talking about CoT.

\subsection{Large Language Models}
Large Language Models (LLM) refer to huge transformer architecture models (typically above ten billions of parameters) pre-trained with massive corpus \cite{zhao2023survey}. With the increment of model size and training corpus, it starts to emerge some new abilities \cite{wei2022emergent}. Recent years, LLM have achieved remarkable progress in many NLP fields \cite{GPT3, palm, flan-palm}.

\subsection{Prompting and In-context-learning}
Prompting is a strategy to better elicit the knowledge and ability acquired during training stage of language models by modifying the input samples in a specific manner \cite{prompt-survey, GPT3,schick-schutze-2021-just}.
In-context-learning is a special designed prompting strategy, which prefixed the query sample with a few example demonstrations including both queries and answers \cite{wei2022emergent}, enabling the model to analogously make predictions \cite{icl-survey}. In-context-learning is a training-free paradigm and can significantly boost LLM performance and data-efficiency across many NLP benchmarks on few-shot scenarios \cite{sun2022blackbox}. 

\subsection{Reasoning on LLM}
Reasoning is a complex process which involves using evidence, logically thinking and making arguments \cite{huang2022reasoning}. It has been a long journey to explore the way to make neural network a reasoning machine \cite{peng2015neural, pmlr-v80-barrett18a, NEURIPS2019_13e5ebb0, DBLP:conf/smc/AngelovS20}. Recently, combined with in-context-learning strategy, LLM showed the prominent progress in reasoning tasks \cite{GPT3, palm, flan-palm}. Especially, with assistance of Chain-of-Thought prompt \cite{CoT}, neural networks made an unprecedented breakthrough on many reasoning benchmarks \cite{selfconsistency, faithful-CoT, fu2023complexitybased}. Some work showed that the reasoning ability may emerge when language models are at a certain scale \cite{cobbe2021training, wei2022emergent, CoT, huang2022reasoning}.

\usetikzlibrary{shadows}
\tikzset{
    my node/.style={
        draw=gray,
        inner color=gray!5,
        outer color=gray!10,
        thick,
        minimum width=1cm,
        rounded corners=3,
        drop shadow,
    }
}
\tikzstyle{bag} = [align=center]
\begin{figure*}[!ht]
\centering
\begin{forest}
    for tree={%
        my node,
        l sep+=2pt,
        grow'=east,
        edge={gray, thick},
        parent anchor=east,
        child anchor=west,
              where level=0{%
                font=\fontsize{7}{7}\selectfont,
                calign = child,
                calign child = 2,
                rounded corners=2pt,
                for descendants={%
                child anchor=east,
                parent anchor=west,
                align=right,
                anchor=west,
                edge path={
                    \noexpand\path [draw, \forestoption{edge}] (!u.parent anchor) -- +(10pt,0) |- (.child anchor)\forestoption{edge label};
                },
              },
              }{where level=1{%
                  font=\fontsize{6}{6}\selectfont,
                  inner color=gray!5,
                  outer color=gray!10,
                  minimum width=1.6cm,
                  rounded corners=2,
                  align=center,
                  parent anchor=east,
                  for descendants={%
                    child anchor=east,
                    parent anchor=west,
                    align=left,
                    anchor=west,
                    edge path={
                      \noexpand\path[\forestoption{edge}]
                      (!u.parent anchor) -- +(10pt,0) |- (.child anchor)\forestoption{edge label};
                    },
                  },
                }{where level=2{%
                  font=\fontsize{6}{6}\selectfont,
                  inner color=gray!5,
                  outer color=gray!10,
                  minimum width=3cm,
                  rounded corners=2,
                  align=center,
                  parent anchor=east,
                  for descendants={%
                    child anchor=east,
                    parent anchor=west,
                    align=left,
                    anchor=west,
                    edge path={
                      \noexpand\path[\forestoption{edge}]
                      (!u.parent anchor) -- +(2pt,0) |- (.child anchor)\forestoption{edge label};
                    },
                  },
                }{where level=3{%
                  font=\fontsize{5}{5}\selectfont,
                  inner color=blue!5,
                  outer color=blue!10,
                  minimum width=9.5cm,
                  rounded corners=2,
                  align=center,
                  parent anchor=east,
                  for descendants={%
                    child anchor=east,
                    parent anchor=west,
                    align=left,
                    anchor=west,
                    edge path={
                      \noexpand\path[\forestoption{edge}]
                      (!u.parent anchor) -- +(2pt,0) |- (.child anchor)\forestoption{edge label};
                    },
                  },
                }{}%
              }%
            }%
        },
    }
    [\rotatebox{90}{Chain-of-Thought Prompting} 
        [Task Types \cref{4}
            [Close Domain Reasoning \\and QA \cref{4.1}
                [Vanilla CoT\cite{CoT}\text{,} Zeroshot-CoT\cite{zeroshot-CoT}\text{,} Synthetic\cite{synthetic}\text{,} Auto-CoT\cite{AutoCoT}\\ ART\cite{art}\text{,} Active-Prompt\cite{active-prompt}\text{,} APE\cite{zhou2023large}\text{,} Faithfull-CoT\cite{faithful-CoT}\\ Explanation Selection\cite{ye2023explanation} Complexity-based prompting \cite{fu2023complexitybased}\text{,} SC\cite{selfconsistency} \\DIVERSE \cite{li2022advance}\text{,} Rationale-augmented \cite{wang2022rationaleaugmented}\text{,} Table-CoT\cite{chen2023large}\text{,} Dater\cite{ye2023large}\\ SOLIS\cite{zhou2022reflection}\text{,} LtM\cite{LtM}\text{,} Decomposed\cite{decomposed}\text{,} PAL\cite{pal} PoT\cite{PoT}\\Selectioninference\cite{selectioninference}\text{,} Algorithmic prompt \cite{zhou2022teaching}\text{,} LP\cite{guo2023learning}\text{,} MoT\cite{li2023mot}\\ SV\cite{weng2023large}\text{,} MathPrompter\cite{imani2023mathprompter}\text{,} Automate-CoT\cite{shum2023automatic}\text{,}PHP\cite{PHP}]]
            [Open Domain Reasoning \\and QA \cref{4.2}
                [Vanilla CoT\cite{CoT}\text{,} Zeroshot-CoT\cite{zeroshot-CoT}\text{,} Auto-CoT\cite{AutoCoT}\text{,} ART\cite{art}\\ Active-Prompt\cite{active-prompt}\text{,} APE\cite{zhou2023large}\text{,} Explanation Selection\cite{ye2023explanation}\\ iCAP\cite{iteratively}\text{,} Multimodal-CoT\cite{zhang2023multimodal}\text{,} Complexity-based prompting \cite{fu2023complexitybased}\\ Self-ask\cite{self-ask}\text{,} SC\cite{selfconsistency}\text{,} DIVERSE \cite{li2022advance}\text{,} SV\cite{weng2023large}\\ Rationale-augmented \cite{wang2022rationaleaugmented}\text{,} IRCoT\cite{trivedi2022interleaving}\text{,} LtM\cite{LtM}\text{,}  MCR\cite{yoran2023answering}\\Decomposed\cite{decomposed}\text{,} Maieutic\cite{maieutic}\text{,} PINTO\cite{wang2023pinto}\text{,} MoT\cite{li2023mot}\\Chameleon\cite{lu2023chameleon} \text{,} Automate-CoT\cite{shum2023automatic}\text{,} RR\cite{he2022rethinking}]]
            [Code Generation \cref{4.3}
                [Self-planning \cite{jiang2023selfplanning}\text{,} XRICL\cite{shi2022xricl}\text{,} DIN-SQL\cite{pourreza2023dinsql}\\Imitation Attack CoT\cite{li2023feasibility}]]]
        [Prompts\\ Design \cref{7}
            [Demonstrations \cref{7.1}
                [Self-ask\cite{self-ask}\text{,} Auto-CoT\cite{AutoCoT}\text{,} Synthetic\cite{synthetic}\text{,}  MMRSelect\cite{fu2023complexitybased}\\ Complexity-based prompting \cite{fu2023complexitybased}\text{,} Active-Prompt\cite{active-prompt}\text{,} Explanation Selection\cite{ye2023explanation}\\ Algorithmic prompt \cite{zhou2022teaching}\text{,} Concise-CoT\cite{madaan2022text}  \text{,} MoT\cite{li2023mot}\\Automate-CoT\cite{shum2023automatic}]]
            [Textual Instructions \cref{7.2}
                [Zeroshot-CoT\cite{zeroshot-CoT}\text{,} APE\cite{zhou2023large}\text{,} Auto-CoT\cite{AutoCoT}]]]
        [Extension \\ Strategies \cref{6}
            [Ensemble \cref{6.1}
                [SC\cite{selfconsistency}\text{,} SV\cite{weng2023large}\text{,} Complexity-based prompting \cite{fu2023complexitybased} \text{,} MCR\cite{yoran2023answering} \\DIVERSE \cite{li2022advance}\text{,} Rationale-augmented \cite{wang2022rationaleaugmented}]]
            [Sub-problems Division \cref{6.2}
                [LtM\cite{LtM}\text{,} Self-ask\cite{self-ask}\text{,} Selectioninference\cite{selectioninference}\text{,} Decomposed\cite{decomposed}\\ HuggingGPT\cite{hugginggpt}\text{,}  Toolformer\cite{toolformer}\text{,} Dater\cite{ye2023large}\text{,} Dynamix-LtM\cite{drozdov2022compositional}\\RR\cite{he2022rethinking}]]
            [External Assistance \cref{6.3}
                [iCAP\cite{iteratively}\text{,} PoT\cite{PoT}\text{,} PAL\cite{pal}\text{,} Decomposed\cite{decomposed}\\ HuggingGPT\cite{hugginggpt}\text{,}  Toolformer\cite{toolformer}\text{,} Faithfull-CoT\cite{faithful-CoT}\text{,} IRCoT\cite{trivedi2022interleaving}\\
                ART\cite{art}\text{,} Synthetic\cite{synthetic}\text{,} Chameleon\cite{lu2023chameleon}]\text{,}RR\cite{he2022rethinking}]
            [Rationalization \cref{6.4}
                [STaR\cite{STaR}\text{,} LP\cite{guo2023learning}\text{,} PINTO\cite{wang2023pinto}\text{,} PHP\cite{PHP}]]]
    ]
\end{forest}
    \caption{Taxonomy of Chain-of-Thought Prompting Strategies}
    \label{taxonomy}
\end{figure*}
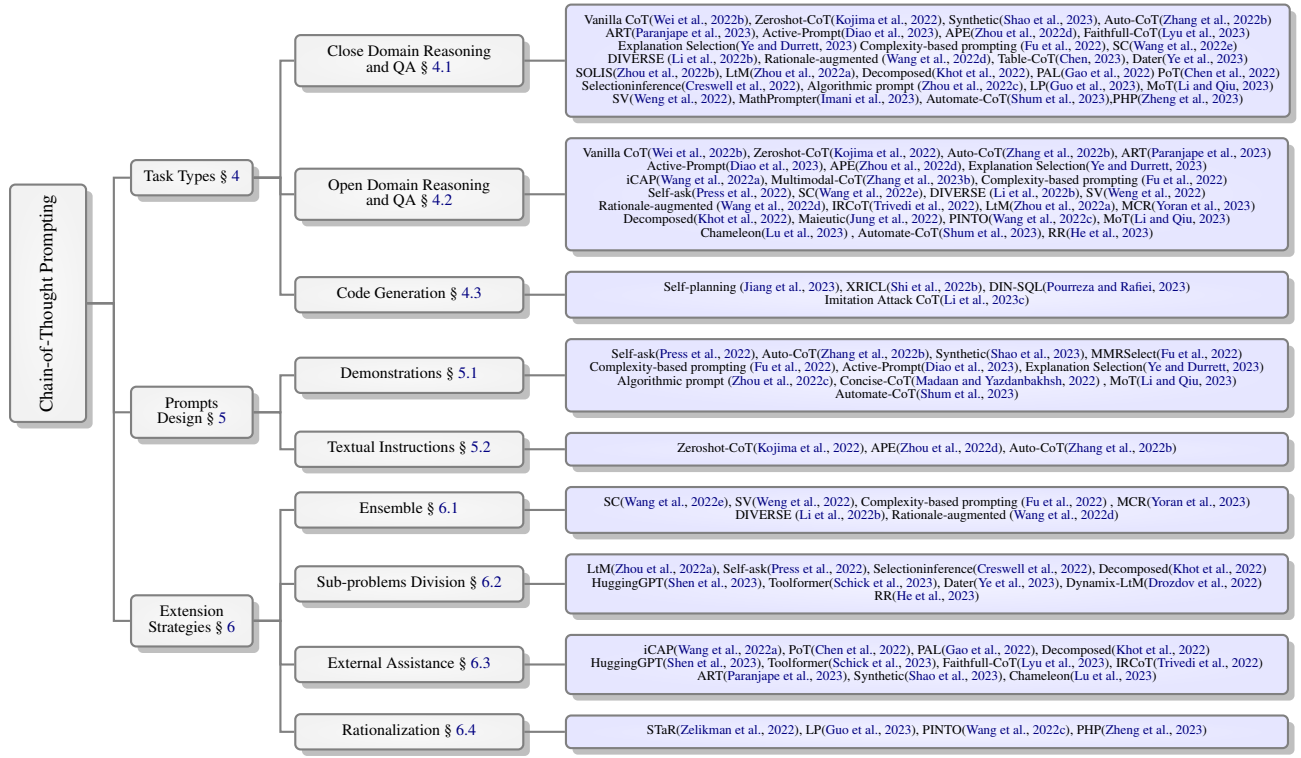

\section{What is Chain-of-Thought Prompting?}
\label{3}
Before formally introducing Chain-of-Thought prompting, we first give a definition of CoT prompts. CoT prompts are special designed input sequences to instruct the model to generate coherent series of intermediate reasoning steps \cite{CoT}. Specially, what refers to the ``\emph{intermediate reasoning steps}"  shows discrepancy in different work. The scope of our survey is a more general covering of present work on prompts for multi-step tasks, which ranges from common step-wise reasoning to multi-step tasks deployment, so we encompass various non-programmatic problems division process including sub-problems decomposition \cite{LtM, self-ask} and multi-step deployment \cite{decomposed, hugginggpt, toolformer} into our definition. 

CoT prompting is a strategy to utilize CoT prompts, which works as following pipeline. As shown in Figure \ref{overview}, given a specific \textbf{task},  CoT \textbf{prompts} are designed on it. With the assistance of optional \textbf{extension strategies}, \textbf{models} predict the answers (typically as well as CoT rationales) based on the input prompts. Above four elements in bold are key factors of CoT prompting pipeline which make significant impact on the final performance. We will separately discuss how these factors influence the prompting effect and introduce the work motivated by these factors in remaining survey. We present the taxonomy of various CoT prompting strategies in Figure \ref{taxonomy}.

\begin{figure*}
\centering
\includegraphics[height=4cm,width=15cm]{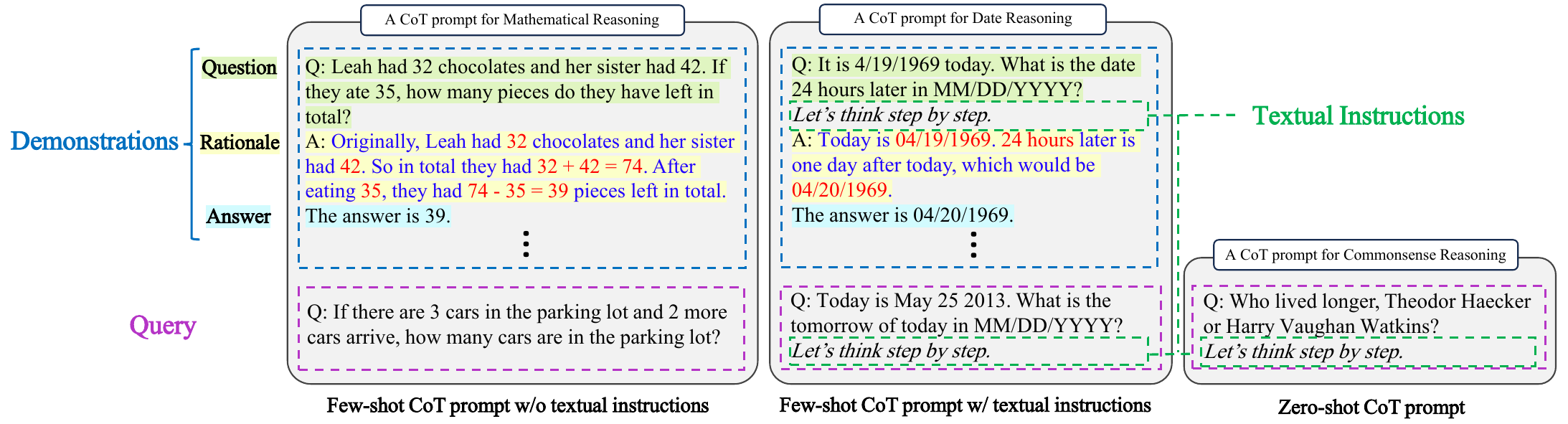}  
\caption{Some example CoT prompts. For rationales in \cref{5.1.2}, \textcolor{red}{\textbf{bridging objects}} are marked in \textcolor{red}{red} and \textcolor{blue}{\textbf{language templates}} are marked in \textcolor{blue}{blue}.}
\label{prompt}
\end{figure*}

\section{Task Types}
\label{4}
Task is the target when utilizing CoT prompting, which makes a foundational difference. Before designing CoT prompting strategies, it is necessary to clarify what types of tasks are prone to being boosted by CoT prompts.

\subsection{Close Domain Reasoning and QA}
\label{4.1}
These kinds of tasks include all necessary conditions and background knowledge in the problems. Models need to select informative materials and conduct reasoning on these materials. CoT prompts could provide a reasoning pattern to instruct how to select key materials and reason on them, showing superiority on these tasks like mathematical reasoning \cite{AutoCoT, synthetic, selfconsistency}, symbolic reasoning \cite{CoT, anil2022exploring, suzgun2022challenging} and table QA \cite{ye2023large, chen2023large}.

\subsection{Open Domain Reasoning and QA}
\label{4.2}
These kinds of tasks aim to answer questions based on a large-scale unstructured knowledge base, and don't include all of necessary knowledge in the problem. In this case, LLM are forced to use their own knowledge to solve the problems and the performance of CoT prompting highly depends on the LLM knowledge quality \cite{suzgun2022challenging}. Some tasks even require a deeper understanding on the semantic meaning of natural language \cite{shi2022language, suzgun2022challenging, zhang2023multimodal}. Improperly involving CoT prompting into these knowledge or semantic oriented tasks may even hurt the performance \cite{shi2022language}. To address these problems, some work uses external tools to inject required knowledge (We will introduce in \cref{6.3}).

\subsection{Code Generation}
\label{4.3}
Code generation aims to generate codes according to an input instruction. Due to the internal logical form of codes, the step-by-step reasoning chain of CoT is coherent with abilities needed for code generation \cite{shi2022xricl, PoT, pal, jiang2023selfplanning, pourreza2023dinsql}. 

\section{Prompts Design}
\label{7}
When the task is determined, it's necessary to design an effective CoT prompt. Besides the query, there are two special elements in CoT prompts: \emph{CoT demonstrations} and \emph{textual instructions}. \emph{CoT demonstrations} are several step-wise reasoning exemplars and \emph{textual instructions} are textual sequences to actively guide a progressive solving process. A CoT prompt should contain at least one of these elements. Typically, a CoT prompt with (or without) demonstrations is called few-shot CoT (or zero-shot CoT). We present several example CoT prompts in Figure \ref{prompt}. 

In this section, we will discuss how to design an effective CoT prompt from each element.

\subsection{Demonstrations}
\label{7.1}
In few-shot CoT, demonstrations are the indispensable part. As shown in Figure \ref{prompt}, a CoT demonstration is a \verb|(problem, rationale, answer)| triple, where the \verb|rationale| contains the intermediate reasoning steps that lead from the \verb|problem| to the \verb|answer|. In this part, we will discuss what factors of demonstrations may affect the final prompting performance from problem, rationale and holistic perspectives\footnote{We don't specially analyze factors of the answer since the answer is directly associated with the problem.}.

\subsubsection{Problem Perspective}
\paragraph{Complexity:} Complexity measures the difficulty of a problem, which can be reflected as the number, the length and the logical difficulty of reasoning steps to solve this problem. 

Empirically, more complex demonstration problems usually guarantee longer reasoning steps, which can provide more reasoning context and induce the model to generate a longer rationale, avoiding a short-cut prediction \cite{synthetic, anil2022exploring, saparov2023language}. To get more complex demonstration problems, a direct way is to select the samples with most reasoning steps \cite{synthetic, fu2023complexitybased}. \newcite{active-prompt} select the demonstrations with higher uncertainty, which represents the extent a problem is able to confuse the model.

\paragraph{Relevance and Diversity:}
Relevance measures how similar the demonstration problems are to the query problem. Relevant demonstration problems could provide similar knowledge and reasoning pattern for the query, which make LLM easier to imitate \cite{ye2022complementary}. 

Diversity measures how different a demonstration problem is from other demonstration problems within a single prompt. Diverse demonstrations could fuse different reasoning process and make model less sensitive to a specific reasoning process, increasing the prompt robustness \cite{synthetic, AutoCoT}.

Both relevance and diversity can be  ensured by representative features of problems \cite{ye2022complementary, AutoCoT} or manual topic informing \cite{synthetic}. Specially, we need to point out that diversity and relevance are a trade-off process \cite{ye2022complementary}. Too monotonous demonstrations may lead to a less robust model performance while involving too many irrelevant demonstrations may inject more noise.  

\subsubsection{Rationale Perspective}
\label{5.1.2}
\paragraph{Structural Completeness:} A CoT rationale can be divided into two fine-grained components\footnote{Besides, \newcite{madaan2022text} decomposed a CoT prompt into a finer granularity: symbols (the tokens which are meant to be reasoned), patterns (the essential or structural tokens which connect symbols) and text (other textual components). This division scheme can be easily inducted into the scheme we introduced.}: \emph{bridging objects} and \emph{language templates} \cite{wang2022understanding, ye2022complementary}. A structural complete rationale should contain both of two components.

\emph{Bridging objects} refer to the critical elements that convey the logical process when solving problems, which are meant to be traversed in a single prediction. \emph{Language templates} refer to complementary textual parts that connect the bridging objects, which involve complementary knowledge for the problems and construct logical connections with bridging objects \cite{wang2022understanding}. Figure \ref{prompt} presents some examples about each component of CoT rationales on different tasks. For a sound reasoning, bridging objects provide the reasoning materials and lead to logical language templates, while language templates provide the essential knowledge and help to better organize bridging objects in back \cite{madaan2022text}. In a word, bridging objects and language templates are two codependent and indispensable components for an effective CoT rationale. More detailed two components can also jointly reduce the ambiguity of demonstration rationales \cite{zhou2022teaching}.

\paragraph{Validity:}Validity measures the correctness and coherence of a reasoning chain. The validity of a CoT rationale can be measured from different metrics like accuracy of answers \cite{NEURIPS2022_c4025018, saparov2023language}, perplexity \cite{self-ask} and more detailed measurement on each reasoning step \cite{saparov2023language}. It's also possible to train a verifier model to evaluate the validity of CoT rationales \cite{li2022advance}. 

Some work is conducted to explore how the validity plays a role in CoT prompt. In a nutshell, a valid CoT rationale does promote a better prompting performance while an effective CoT prompt does not necessarily demand a totally valid rationale \cite{wang2022understanding, madaan2022text, ye2022complementary, chen2023demonstrations}. \newcite{wang2022understanding} found that a completely wrong rationale may bring a similar reasoning boost as vanilla CoT as long as it is coherent, \emph{i.e.}, latter steps should not contain any condition or text information which doesn't present in previous steps. It seems that a coherent CoT rationale demonstrates a relative logical manner of reasoning, which help to elicit the LLM ability of progressive reasoning. Thus, the coherence should be primarily ensured if we have to compromise to incorrect CoT rationales.

\subsubsection{Holistic Perspective}
\paragraph{Number and Order:} When prompted with multiple CoT demonstrations, the model accesses the intersection information of each demonstration \cite{ye2022complementary}. The number and order of demonstrations leave a impact on the final performance \cite{CoT, NEURIPS2022_11332b6b, chen2023demonstrations}

Some work finds model performance would gain prominently when the CoT demonstrations number is gradually increasing from zero to two while the improvement remains slowly and negligibly if demonstrations number continually increases \cite{shi2022language, NEURIPS2022_11332b6b, chen2023large, chen2023demonstrations}. Too many demonstrations can lead to a mass of computation cost while insufficient demonstrations may make LLM more sensitive to a single demonstration and increase the variance \cite{PoT}.

Changing the order of demonstrations may result in a non-trivial effect, but we still can't draw a conclusion for a universal order strategy since the impact of order may vary according to the models, tasks and dataset \cite{lu-etal-2022-fantastically, liu-etal-2022-makes}. An compromised method is to use some prompting searching strategies \cite{lu-etal-2022-fantastically, ye2023explanation} or some heuristic measurements like complexity and relevance order \newcite{liu-etal-2022-makes}.

\subsection{Textual Instructions}
\label{7.2}
LLM show ability to follow explicit instructions even in zero-shot scenarios \cite{ouyang2022training, sanh2022multitask}. Inspired by this, some work finds explicitly prompting LLM with an active textual instruction like ``\emph{Let's think step by step}"  can guide a progressive reasoning \cite{zeroshot-CoT, zhou2023large}. Without any demonstrations, this simple zero-shot strategy shows impressive result comparing to non-CoT methods, implying these textual instructions can similarly elicit the reasoning ability of LLM. Some work also finds combining these textual instruction with few-shot CoT can achieve a further performance increment \cite{zeroshot-CoT}. 

\section{Extension Strategies}
\label{6}
Given a CoT prompt, there are many possible extension strategies to enhance the prompt performance. In this section, we will highlight four CoT-related strategies and analyze when and how to use them.

\subsection{Ensemble}
\label{6.1}
Ensemble learning is an effective strategy which combines diverse learners, enhancing the model performance comparing to a single learner \cite{zhou2012ensemble}. Recent work achieved superior performance when using ensemble strategy on CoT prompting \cite{wang2022rationaleaugmented, li2022advance, selfconsistency}, which can help to correct errors made by individual reasoning process and integrate diverse prompts and demonstrations into a single prediction. \newcite{selfconsistency} point out ensemble methods can even bring performance increment on tasks where vanilla CoT fails. However, unnecessary ensembles on problems which vanilla CoT can already effectively solve may inject noise to a confident prediction and instead do harm to model performance \cite{wang2022rationaleaugmented}.

To go a step further, what elements should be embraced into ensemble also matters a lot. According to different ensemble materials, we categorize these methods into \emph{prompts ensemble} method and \emph{predictions ensemble} method. 
\emph{Prompts ensemble} focuses on the ensemble of results generated with various prompts. This method construct diverse CoT prompts by repeating sampling different demonstrations from exemplars set \cite{li2022advance}.
\emph{Predictions ensemble} focuses on integrating output space materials including rationales and answers. This method generates various predictions given a fixed input query by LLM sampling algorithms \cite{selfconsistency, fu2023complexitybased, yoran2023answering}. It is found predictions ensemble may lead to more performance gain comparing to prompts ensemble \cite{wang2022rationaleaugmented}, but multiple decoding for predictions ensemble may lead to higher computation cost. How to choose the ensemble strategy depends on the access of demonstrations number and computing resource. It's also possible to jointly combine two ensemble methods \cite{wang2022rationaleaugmented}.

\subsection{Sub-problems Division}
\label{6.2}
When confronting a problem needs to be recursively inferred or harder than demonstrations, dividing a problem into several sub-problems could be a better option \cite{LtM, self-ask, decomposed, toolformer}. Comparing to vanilla CoT, sub-problems division strategy decomposes a complex problem into a series of simple sub-problems, which are much easier to solve, enabling models to accomplish query problems harder than demonstrations \cite{LtM, self-ask}. Also, when dealing with each sub-problem, model is free from information which is irrelevant to current sub-problem and more informative information is prone to guiding a valid reasoning \cite{selectioninference, LtM}. Additionally, the required abilities for separate sub-problems are different. This strategy makes it more convenient to deploy each sub-problem with different modules and inject external assistance \cite{decomposed, toolformer, hugginggpt}, which we will introduce in \cref{6.3}.

\subsection{External Assistance} 
\label{6.3}
In order to expand the ability of LLM and assist LLM to perform on broader applications, it's useful to introduce external sources including knowledge, tools or codes interpreters into reasoning process. 

Knowledge injection is especially helpful in tasks which need external knowledge like commonsense QA \cite{iteratively}. Tools and codes assisted strategies show preponderance in problems which need abilities beyond LLM capacities such as accurate numerical calculation or search engine \cite{PoT, decomposed, toolformer, art}. With proper prompts, LLM can generate task deployment chains to instruct when and where to call external tools \cite{decomposed, toolformer}, codes interpreter \cite{pal, PoT, faithful-CoT} or even other models \cite{hugginggpt}, to solve more complex problems.

Codes interpreters can also serve as external verifiers to check the validity of generated rationales by checking whether they can be interpreted and lead to a correct answer \cite{synthetic, pal, PoT}. Additionally, \newcite{faithful-CoT} point out executing reasoning chains with programmatic modules can enhance the faithfulness of CoT (We will discuss in \cref{8}).

\subsection{Rationalization} 
\label{6.4}
Usually, the rationales predicted by LLM would make some mistakes and lead to wrong answers. If these mistakes can be corrected, it is possible to rationalize the reasoning process and boost the performance. 

Manual rationalization would be effective but sometimes too costly \cite{wang2023pinto, kim2023cotever}. A simple way is to use some hints to guide the model to rethink \cite{STaR, guo2023learning}. When the model produces a wrong answer, we can tell model the correct answer and ask it to self-revise illogical reasoning and regenerate a rationale based on the golden answer. This process can be regarded as a self-learning process, where the model can progressively improve its reasoning ability just with answers supervised. However, it's still hard to rationalize imperfect rationales which lead to correct answers.

\section{Models}
\label{5}
LLM, as the primary role of solving problems, makes a significant difference to the final prediction. In this section, we will discuss from model size and training corpus to introduce what kinds of models are more effective with CoT prompting. 

\subsection{Model Size}
Many researches have found as the model size is relatively small (typically below ten billions parameters), CoT doesn't remain a positive impact. But as the model size increases to a certain size (above ten billions parameters), it will exhibit a sudden performance breakout \cite{CoT, suzgun2022challenging, magister2022teaching, fu2023specializing}. This implies CoT is an emergent ability \cite{wei2022emergent} of LLM. Prompting small models with CoT will commonly lead to hallucination \cite{hallucination}, which often presents as fluent but illogical generation \cite{CoT}. 

But it's still possible to enhance small models reasoning ability by CoT. Some work fine-tuned a small-scaled model with self-constructed CoT dataset\cite{STaR} or knowledge distillation \cite{magister2022teaching, ho2022large, wang2023pinto, fu2023specializing}, making small models compatible to perform step-by-step reasoning even on few-shot scenarios.  However, small models will forget general abilities on other tasks except step-by-step reasoning after CoT tuning \cite{fu2023specializing} and still lag behind large models on tasks which demands substantial knowledge to conduct reasoning \cite{magister2022teaching}. 

\subsection{Training Corpus}
It is believed that the abilities LLM exhibit originate from training corpus. Some work finds models pre-trained with codes could acquire more performance gain when prompted with CoT \cite{active-prompt}. Instruction tuning also shows relevance to the CoT prompting and zero-shot learning performance\cite{flan-palm, fu2023specializing}, which may be illustrated by the presence of CoT-like samples in the training corpus of instruction tuning. Recent work even tries to explicitly involve CoT samples into training corpus to enhance the step-by-step reasoning ability and avoid over-fitting to monotonous sample templates \cite{flan-palm, ho2022large, yu2022alert}. In a word, embracing aforementioned contents into training corpus could introduce more reasoning materials and necessary knowledge for LLM, leading to a profound influence on CoT reasoning ability. 

\section{Discussion and Future Work}
\label{8}
\paragraph{Faithfulness:} Though CoT prompting increases the reasoning interpretability, it still remains a problem whether present methods for generating CoT rationales are faithful. A faithful reasoning process implies the answer generated by model is accurately reasoned from corresponding generated rationale, which can ensure the controllability and credibility of reasoning process. The rationale which looks like plausible but unfaithful may make humans over-trust the model, resulting in implicit bias risk on realistic applications \cite{pruthi-etal-2020-learning, 10.1145/3375627.3375830}. 

Most of current work generates CoT rationales together with the final answer, which can't guarantee the answer is directly acquired from rationales. Although there exists some work that tries to construct a faithful process by directly reasoning answers on generated rationales \cite{faithful-CoT, wang2023pinto}, it still can't ensure the faithfulness about how rationales generated from the query problem \cite{faithful-CoT}. How to perform a truly faithful step-by-step reasoning on LLM can be critical for controllable and reliable CoT reasoning.

\paragraph{Generality:} Currently, it has still a challenge for CoT-assisted LLM to handle the problems which need a great amount of external knowledge or deeper understanding of language \cite{shi2022language, suzgun2022challenging, zhang2023multimodal}. Much relevant work focus on an external retriever to supplement necessary knowledge for conduct reasoning. However, a pre-trained LLM has already obtained a tremendous amount of knowledge. The key conundrum is how to instruct LLM to recognize and make use of learned knowledge. 

Besides, although LLM with CoT assistance have already showed impressive advantages on some reasoning tasks, these tasks are a little bit na\"ive and still far from complicated realistic applications, which demand a high level of perception on external environments, deduction of potential results and planning on the final goal even with cost restriction. \newcite{valmeekam2023large} proposed a more complicated planning benchmark which evaluates the model reasoning ability about actions execution and environment change, finding that LLM have dismal performance on this benchmark. How to combine the CoT prompting with LLM on more challenging and realistic benchmarks can be a promising future direction.

\paragraph{Self-rationalization:} Recent work has proposed effective self-rationalization methods by informing the model correct answers and ask models to self-revise the mistakes\cite{STaR, guo2023learning}. But sometimes models may learn some short-cut and generate correct answers while incorrect rationales \cite{saparov2023language}. In this scenario, answer accuracy is not enough to measure the validity of reasoning process. How to spot these short-cut and help models to self-rationalize incorrect reasoning should be carefully discussed.

\paragraph{Rationale Analysis:}  Existing work has made a step towards a clear understanding of CoT rationale, but most existing conclusions are confined to limited models and tasks \cite{wang2022understanding, ye2022complementary, madaan2022text}. The specific form of CoT rationale varies according to the specific task setting. How a certain component is represented and how each component behaves on the given setting may leave a great distinction. Therefore, for a more practical application, it's vital to thoroughly analyze the function of each rationale component among different settings. 

As we discussed in \cref{7.1}, in order to achieve a promising prompt performance, a relatively complex and detailed reasoning chain is needed. With this purpose, the input tokens are usually long, increasing the inference cost and noise interference. \newcite{madaan2022text} proposed Concise-CoT to prune a few trivial tokens but there is still a long journey to go. If we could fully understand how each component work for each task, we could directly prompt partial demanded reasoning materials instead of a complete reasoning chain to improve the efficiency of CoT prompting.

\paragraph{Theoretical Analysis:} We have already showed CoT prompting can benefit the reasoning performance from different dimensions. But it naturally leaves a question about the theoretical explanation about CoT prompt effect. 

There are some hypotheses to explain the mechanism of in-context-learning \cite{chan2022data, xie2022an, olsson2022incontext, garg2023transformers, li2023transformers, dai2023gpt}. But CoT prompts are typically more complicated due to the step-by-step reasoning chains, thus there may exist more to explore for CoT prompting. Unfortunately, there isn't enough theoretical analysis on CoT prompting and more work draws empirical conclusions. Some work finds prompting with wrong and irrelevant rationales can somewhat boost the reasoning performance of LLM \cite{wang2022understanding, madaan2022text}. This may imply CoT prompts don't ``teach" LLM how to reason, but instead provide necessary reasoning materials to help LLM to recall what they have learned at training stage. \newcite{prystawski2023think} give a hypothesis from training data locality structure and \newcite{li2023dissecting} provide an explanation from learning perspective. But how these hypotheses can explain aforementioned demonstration factors we emphasized is still a mystery. Also, we still don't know how a simple ``\emph{Let's think step by step}" help to boost the performance. Clarifying the working mechanism of CoT prompting is a fundamental milestone to ensure a truly interpretable and transparent prompting strategy. 

\section{Conclusion}
In this survey, we reviewed the research status of Chain-of-Thought prompting. We highlighted four factors that may affect the CoT prompting performance and introduced methods based on these factors. We gave a general direction to properly utilize CoT prompting when confronting different setting. Furthermore, we discussed current challenges about CoT prompting and proposed some potential directions. We hope this survey could provide an overall reference on future research.

\section*{Limitations}
Due to the page limitation, we couldn't concretely introduce the methods designed on various tasks and applications setting, but only provide an systematic analysis and instructive perspective on prompting applications. We don't specifically introduce the work which tries to use CoT prompting for more realistic applications such as examinations problems answering \cite{zhang2022automatically}, domain specialization \cite{singhal2022large, DBLP:journals/corr/abs-2209-08141, zhang2023causal}, since these work doesn't exceed the scope we discussed in \cref{4}. There are also other applications for CoT on imitation learning and reinforced learning \cite{yang2022chain, jia2023chainofthought, bai2022constitutional}. We don't discuss in out paper because we only focus on prompting applications of CoT.

Besides, since the scope of our survey focuses on the factors analysis and strategies designing of CoT prompting, we don't encompass application resources in main text (refer to Appendix \ref{B}).



\bibliography{ref}
\bibliographystyle{acl_natbib}

\appendix
\section{Related Surveys}
\label{A}
Some recent surveys on prompts and in-context-learning contain the introduction of Chain-of-Thought prompting but most of them only briefly introduce some relevant methods and lack of a systematic and comprehensive analysis. \newcite{daull2023complex} and \newcite{yu2023natural} focused on techniques on QA and reasoning tasks and just contain a brief mention on CoT prompting. \newcite{zhao2023survey} introduced Chain-of-Thought prompting on few-shot and zero-shot scenarios but did not make a deep analysis on CoT prompting strategies designing. \newcite{huang2022reasoning} and \newcite{icl-survey} introduced common methods of CoT prompting, but they did not give a detailed formalization on CoT prompt and a systematic taxonomy of these methods. These work also didn't cover comprehensive work of CoT prompting. Closer to our work, \newcite{qiao2023reasoning} presented a survey on prompts for reasoning tasks and introduced some work on CoT prompting. Comparing to their work, we focus more on the deep analysis when utilizing CoT prompting. For example, when it comes to task applications, \newcite{qiao2023reasoning} just introduce the form of different tasks while we discussed the characteristic of these tasks and explained why and how to use CoT prompting on these tasks. We are also the only survey which contains a fine-grained formulation of CoT prompt.

In short, unlike methods collection surveys, our survey aims to provide a deeper and more comprehensive analysis on CoT prompting. We want to give a general guide for communities to better utilize the CoT prompting and provide a clear vision on prompting strategies designing.

\section{Relevant Resources}
\label{B}
Since CoT shows prominent strength in many tasks involving multi-steps, the acquisition of step-wise data is a key aspect for CoT application. Considering the great manual efforts demand for CoT data annotation, a comprehensive dataset and efficient annotation tool is of great importance. We will introduce some CoT resources in this section.

\newcite{saparov2023language} presented a ontology structure based CoT logical reasoning dataset, PRONTOQA. \newcite{ott2023thoughtsource} proposed a CoT reasoning meta-dataset, ThoughtSource, which integrates six scientific/medical , three general-domain and five math word question answering datasets with manual or AI-generated CoT annotated. The dataset embraces various types of question answering including number, multiple choice, text, bool, \emph{etc}. This dataset can be effectively used in both few-shot in-context-learning and fine-tuning paradigm.

For more efficient CoT data annotation, \newcite{kim2023cotever} proposed a CoT annotation toolkit, CoTEVer, enabling a more applicable manner to reduce manual efforts. CoTEVer can retrieve evidence for generate CoT explanations and provide a more efficient way to verify and correct the wrong explanations.
\end{document}